# Using Genetic Algorithm in the Evolutionary Design of Sequential Logic Circuits

**Parisa Soleimani [1], Reza Sabbaghi-Nadooshan [2] , Sattar Mirzakuchaki[3] , and Mahdi Bagheri [4]**

[1] Member of Scientific Association of Electrical and Electronic Engineering,
Islamic Azad University, Central Tehran branch
Punak, Tehran, Iran
parisa.soleimani@gmail.com

[2] Department of electronic engineering
Islamic Azad University, Central Tehran branch
Punak, Tehran, Iran
r_sabbaghi@iauctb.ac.ir

[3] Department of electronic engineering
Iran University of Science and Technology
Narmak, Tehran, 16846-13114, Iran
M_kuchaki@iust.ac.ir

[4] Department of electronic engineering
Iran University of Science and Technology
Narmak, Tehran, 16846-13114, Iran
m-bagheri@elec.iust.ac.ir

**Abstract**

Evolvable hardware (EHW) is a set of techniques that are based on the idea of combining reconfiguration hardware systems with evolutionary algorithms. In other word, EHW has two sections; the reconfigurable hardware and evolutionary algorithm where the configurations are under the control of an evolutionary algorithm. This paper, suggests a method to design and optimize the synchronous sequential circuits. Genetic algorithm (GA) was applied as evolutionary algorithm. In this approach, for building input combinational logic circuit of each DFF, and also output combinational logic circuit, the cell arrays have been used. The obtained results show that our method can reduce the average number of generations by limitation the search space.

## *1. Introduction*

The aim of evolvable hardware is the self-sufficient reconfiguration of hardware structure in order to improve performance. In designing and optimizing of the evolutionary circuit, an optimization algorithm searches the all space of possible circuits and determines solution circuits with desired functional response. Simpler structure of combinational circuits in compare with sequential circuits and the lack of feedback in this circuits is caused more researches have been done in this field. Different evolutionary algorithms have been used to evolve combinational logic circuits, for example Vasicek used Cartesian genetic programming [1], Stomeo employed evolutionary strategy [2], and Jackson used genetic programming [3].

On the other hand, relatively few efforts have been done to evolve the sequential logic circuits [4]. For example, Higuchi used GA to search for circuits that represent the desired state transition function [5]. Manovit synthesized frequency detector, odd parity detector, module-5 counter, serial adder [6]. Aporntewan evolved serial adder, 0101 detector, module-5 counter, Reversible 8-counter with genetic algorithm [7]. Solimon designed 3-bit up-counter [8], and Shanthi evolved module-6 counter, 'lion' circuit [9].

In this paper, we have proposed a method for designing and optimizing the synchronous sequential logic circuits with 100% functionality and minimal number of logic gates.

In the rest of this paper, sections 2 consider the main idea of the proposed method. Section 3 describes GA operators. Section 4 describes details of process to define structure of chromosomes. Section 5 explains fitness evaluation process to evaluate the performance of evolved circuits. Simulation environment has been described in section 6. Section 7 summarizes the experiment of proposed method on two sequential circuits and shows the simulation results for target circuits. Finally, in section 8 the conclusion of this paper is presented.

## 2. *The Proposed Method*

The structure of sequential logic circuits comprises a set of two sections of combinational logic circuit and D flip-flops [10]. In this approach, for designing combinational parts, we present a constant structure of two dimensional rectangular arrays of logic gates. We put this array to input of each DFF for building their next states, and before the primary outputs to build the outputs of target circuit as Fig. 1. With evaluation of each array separately, speed of evolution is increased and the evolution time is decreased. The described array for building combinational logic parts is shown in Fig. 2. This array has R rows and C columns and their logic gates are chosen from AND, OR, XOR, and NOT gates. Except NOT gate, the other gates have two inputs and one output. Each gate input can be obtained from primary inputs, Present states of DFFs, or output of each left neighbor gate.

One Multiplexer is added to the inputs of gates in each array, the input of DFFs, and before the primary outputs. We change connection between gates and DFFs by changing the selection bits of multiplexers. Hence, by determining the proposed structure of chromosome encoding (section 4) and by using genetic algorithm, we have evaluated the different states of logic gate connections to achieve correct functionality and minimum number of logic gates.

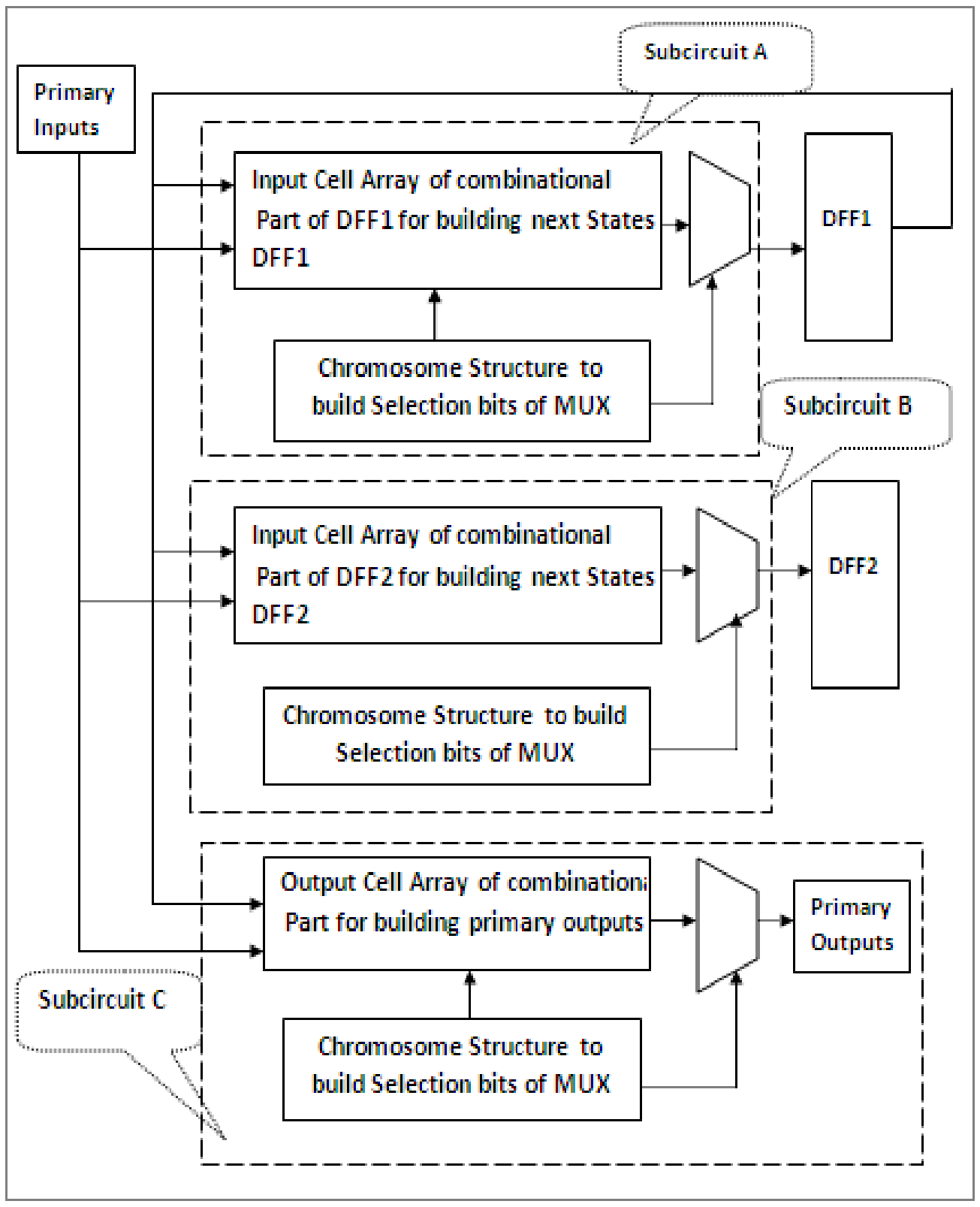

Figure 1. Block diagram of the proposed method for sequential logic circuit with two DFFs

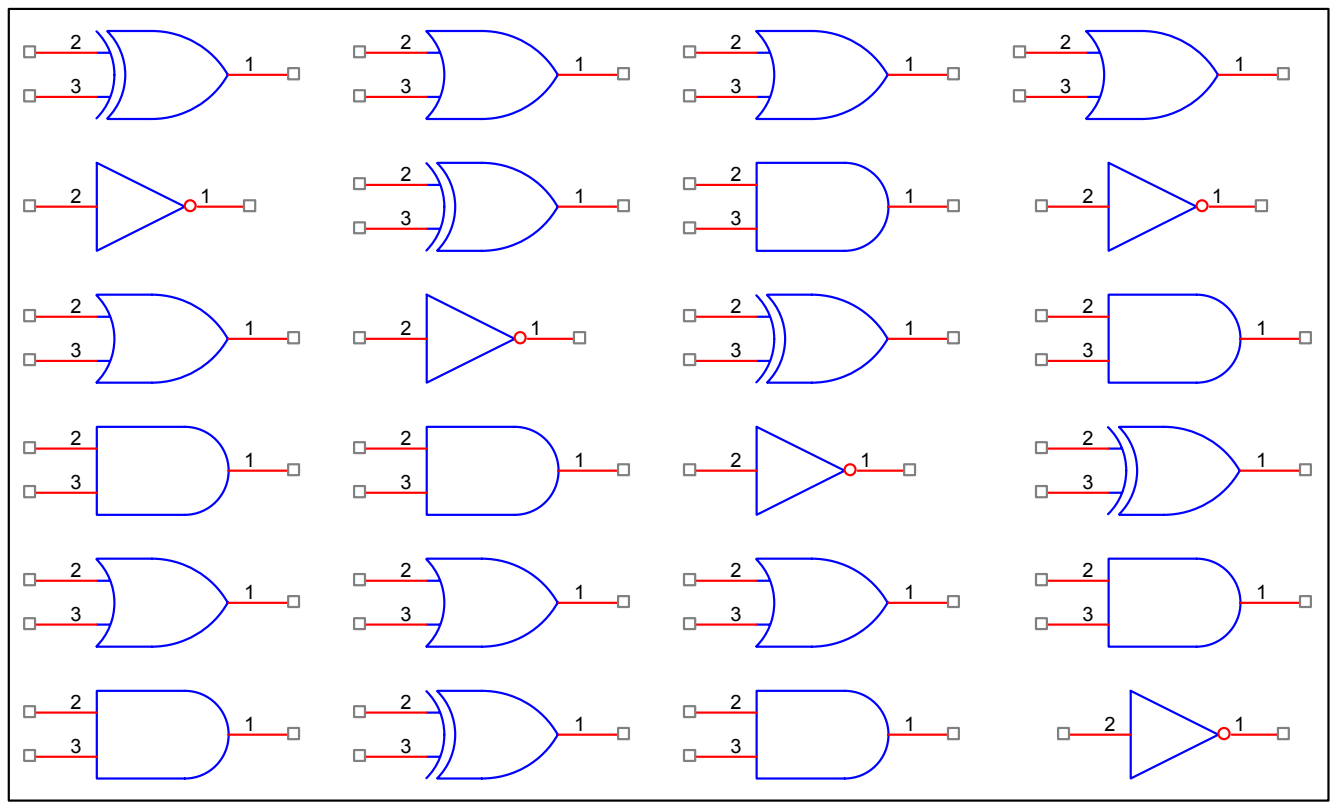

Figure 2. Schematic of the rectangular array structure for building combinational logic parts.

## 3. *Genetic Algorithm Operations*

In this paper, genetic algorithm has been used to evolve the particular circuit. Genetic algorithm is a general search technique that can be applied to search problems where the solution can not be identified within a finite period of time.
In this approach, individuals have been defined in type of bit string. We described genetic algorithm operators as follows:

1. Selection: we have chosen the roulette wheel as a method for parent selection.

2. Crossover: a pair of parents produce child by using one-point crossover.

3. Mutation: mutation is described as a random change of genes in the chromosome. The mutation method that has been used in this study is the uniform mutation. In our experiments, population size has been defined as 10 and maximum number of generations is set to 40,000. The algorithm is stopped if there is no advance in the fitness function for 20,000 consecutive generations, or fitness value reach to -100(maximum value of fitness). This is for overcome the stalling effect.

## *4. Chromosome Encoding*

The basic concept behind the combination of reconfigurable hardware systems and evolutionary algorithm (similar to GAs in EHW) is to regard the configuration bits for the reconfigurable hardware devices as chromosomes for the genetic algorithms. If the fitness function is correctly designed for a task, then the genetic algorithm can autonomously find the best hardware configuration in terms of the chromosomes (i.e. configuration bits).
The chromosome defines the construction of the logic circuit and the connectivity between logic gates. In this approach, we have put a multiplexer to input of each gate, DFFs, and before the primary outputs. Fig. 3 shows block diagram of cell array after adding multiplexer to it.
We changed connection between gates and DFFs by changing the selection bits of multiplexers. Inputs of multiplexers of logic gates are taken from primary inputs, present states of DFFs, outputs of all gates that is the neighbor left column, and constant values that set equal '0' and '1'. Also inputs of multiplexers of DFFs and primary outputs are obtained from primary inputs and outputs of all logic gates that are on the all left columns. Fig. 4 depicts the structure of multiplexer that is used.

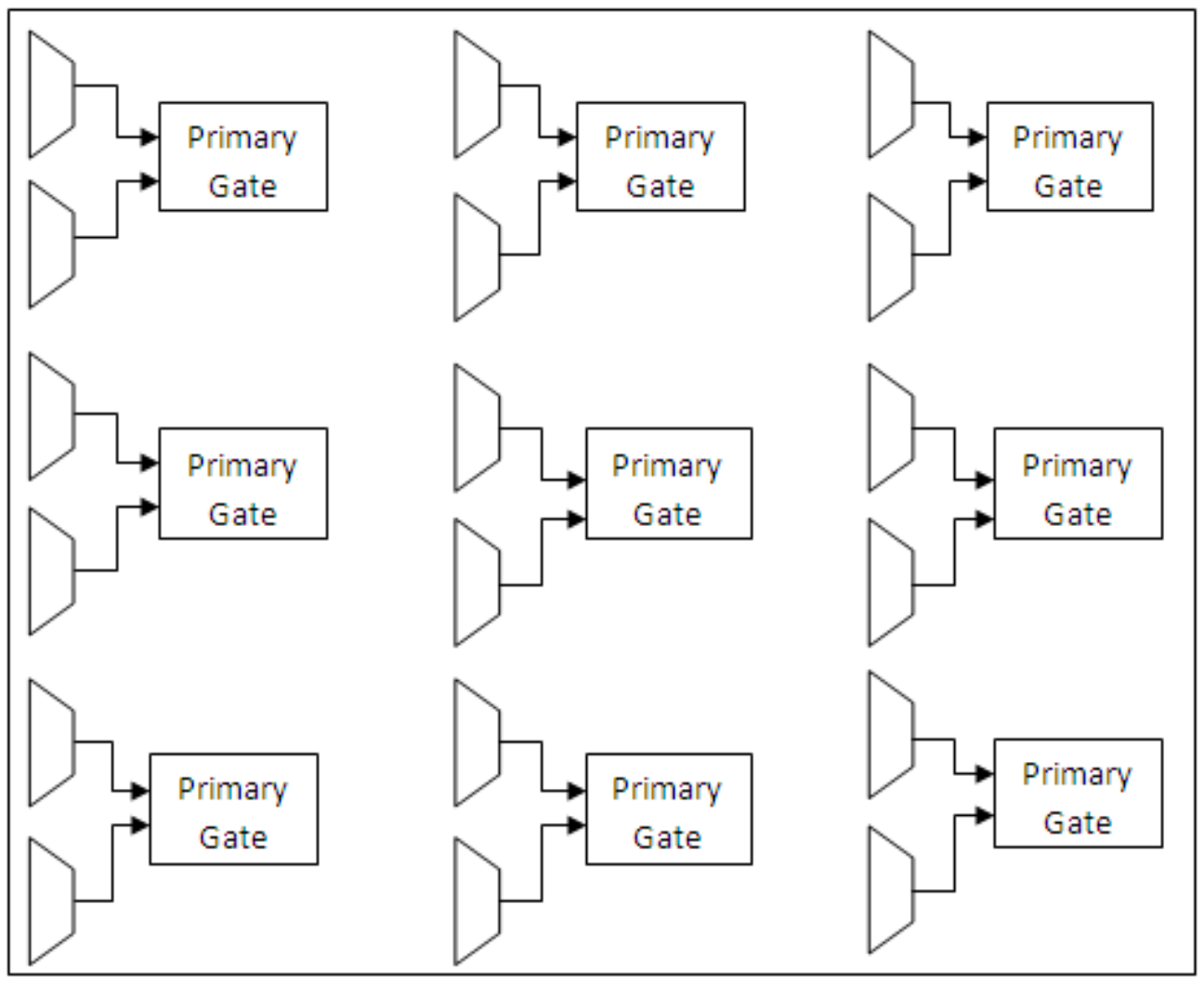

Figure 3. Block diagrom of cell array after adding multiplexer to it.

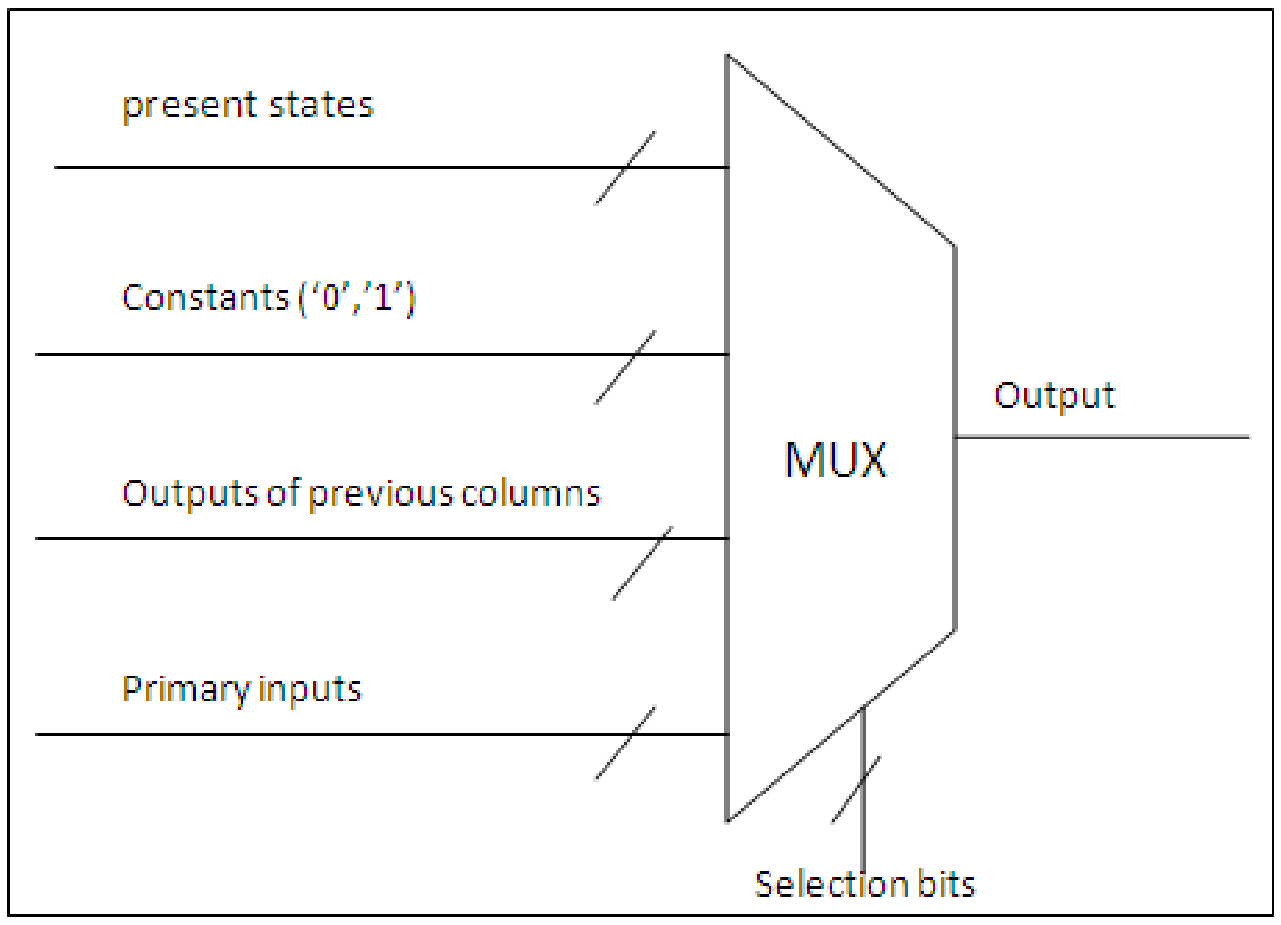

Figure 4. Structure of Multiplexer has been used in "Fig.3"

Changing selection bits of multiplexers leads to different connectivity between logic gates of circuit. We have used the selection bits of multiplexer as chromosome genes as Fig. 5.

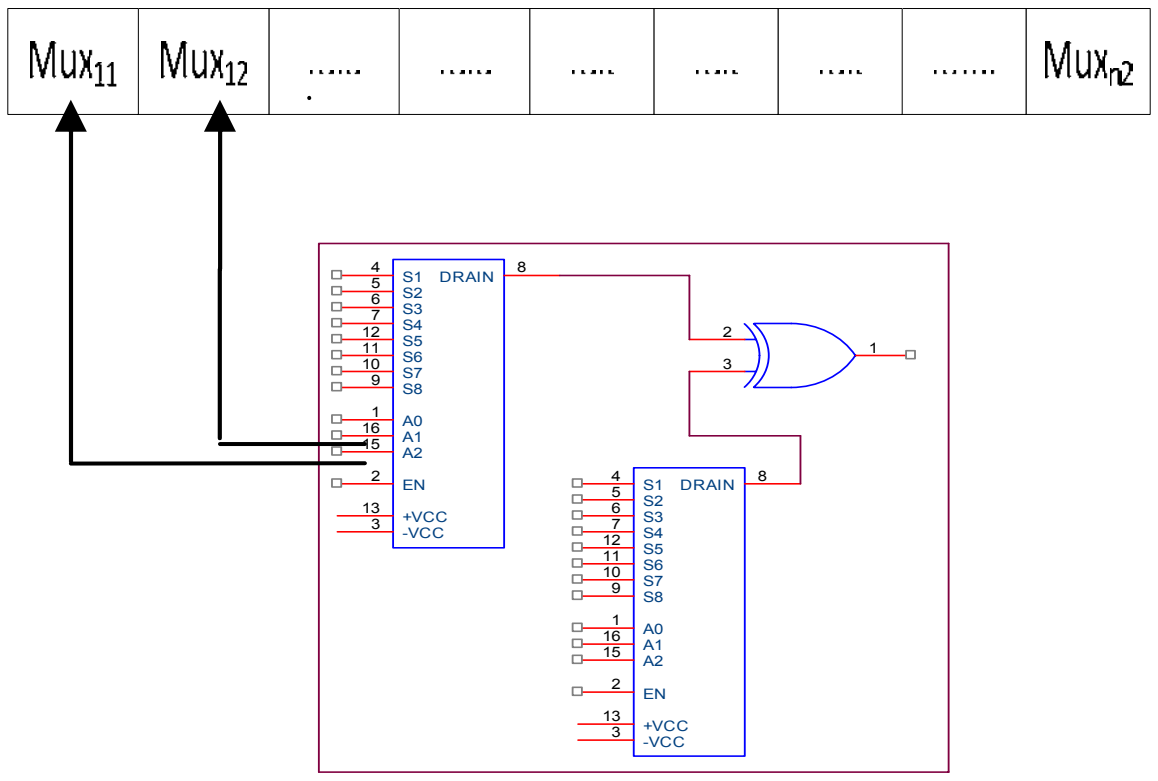

Figure 5. Structure of chromosome encoding.

### 5. *Fitness Evaluation Process*

A fitness function in GA measures goodness of every individual in population with respect to the problem under consideration. We used finite state machine (FSM) for evaluation of sequential circuits. In this method, first the desired state is set in the circuit flip flops and then we changed the value in primary inputs and compared the output of circuit with the desired ones. If these two values are equal, then the fitness value is increased. In proposed method, we measured fitness function by two main criteria: design and optimization. In the first criteria, functionality of the circuit is evaluated. Our first objective indicator is evolving a circuit that has 100% functionality. Then in the second criteria, optimization has been performed by reducing the numbers of logic gates that are used in the target circuit. Fitness optimization is activated once design fitness value reaches 100% functionality.

The design criterion of any individual is evaluated as these steps:

1. The initial value for design fitness has been considered to zero.

2. The primary inputs and present state of DFFs have been set externally. Then the value of next state of DFFs and primary output of the circuit is measured after sending a clock signal to DFF.

3. The corresponding output with desired output has been compared. We can use this equation to measure fitness:

$F_{Design}=F_{Design}+$ number of equal output bits. (1)

4. The steps 2-3 have been repeated for the remaining states of FSM and functionality of circuit has been evaluated.

The optimization criterion has been calculated as follow steps:

1. The initial value for optimization criterion has been considered as:

$F_{Optimization} = R*C$ (2)

2. For each individual, total number of logic gates have been calculated. So, we can use this equation to find optimization fitness:

$F_{Optimization} = (R*C)$ - number of logic gates that is used in new circuit. (3)

Now, the final fitness of individual could be calculated by using this equation:

$F_{final} = F_{Design}+ F_{Optimization}$ (4)

Both of the procedures described above are applied for evaluation of combinational parts of sequential logic circuit.

### 6. *Simulation Environment*

In this method, we used Modelsim as VHDL hardware programming language simulator and MATLAB software for implement GA. Also we used GA toolbox in MATLAB Revision 2010 software to run the evolutionary algorithm. In addition we used simulator link $^{TM}$ MQ toolbox in this software. It can access to Modelsim, open HDL code, run it for different inputs that are determined in MATLAB code and save outputs in the variables of MATLAB codes. Hence this toolbox is as a link between Modelsim and MATLAB. Fig. 6 shows block diagram of this process.

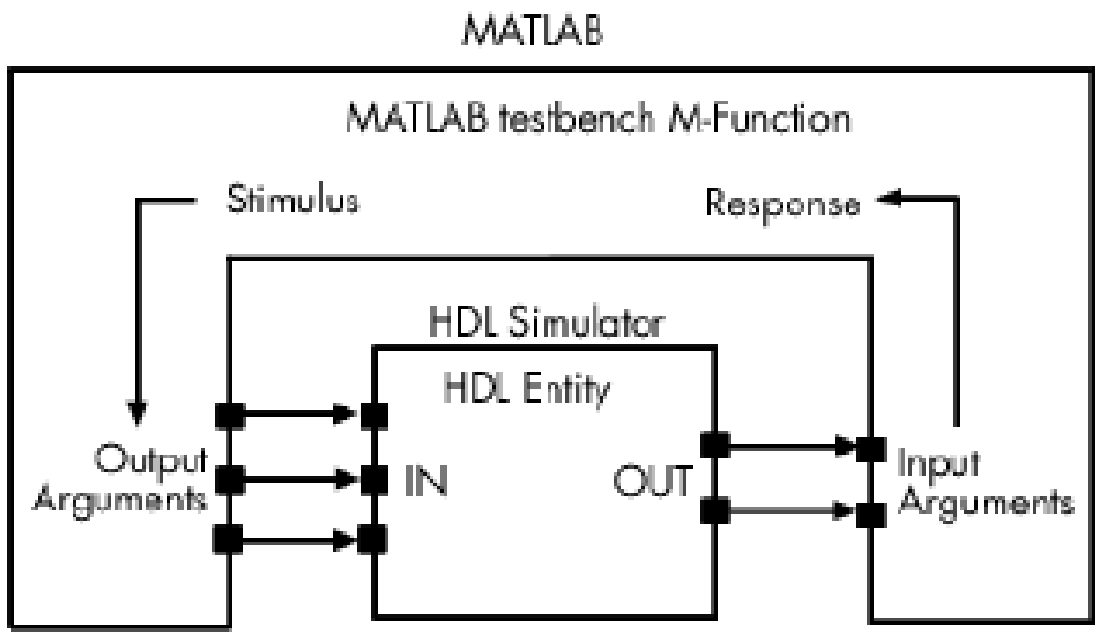

Figure 6. Structure of chromosome encoding[11].

## *7. Experiments and Results*

In this section, the proposed method is experimented on two types of the sequential circuits.

### *a)1010 Sequetial detector*

The first circuit is a 1010 sequential detector. The target sequence detector circuit has one input, one output, and four internal states. State transition graph of this circuit has been shown in Fig. 7. In this method, we designed the target detector based on the symbolic transition table shown in Fig. 8. In this figure, step1 shows the symbolic state table of FSM and state assignment to each state. In step2, STT of the target circuit is shown. In step3, STT of the circuit is divided into input combinational logic subcircuit A and B and output combinational logic subcircuit C [10]. This circuit has four states that uses two DFFs. As we explained in previous sections, we evaluated each subcircuit A, B, and C separately. Finally, the sequential circuit is assembled. The evolved circuit is shown in Fig. 9.

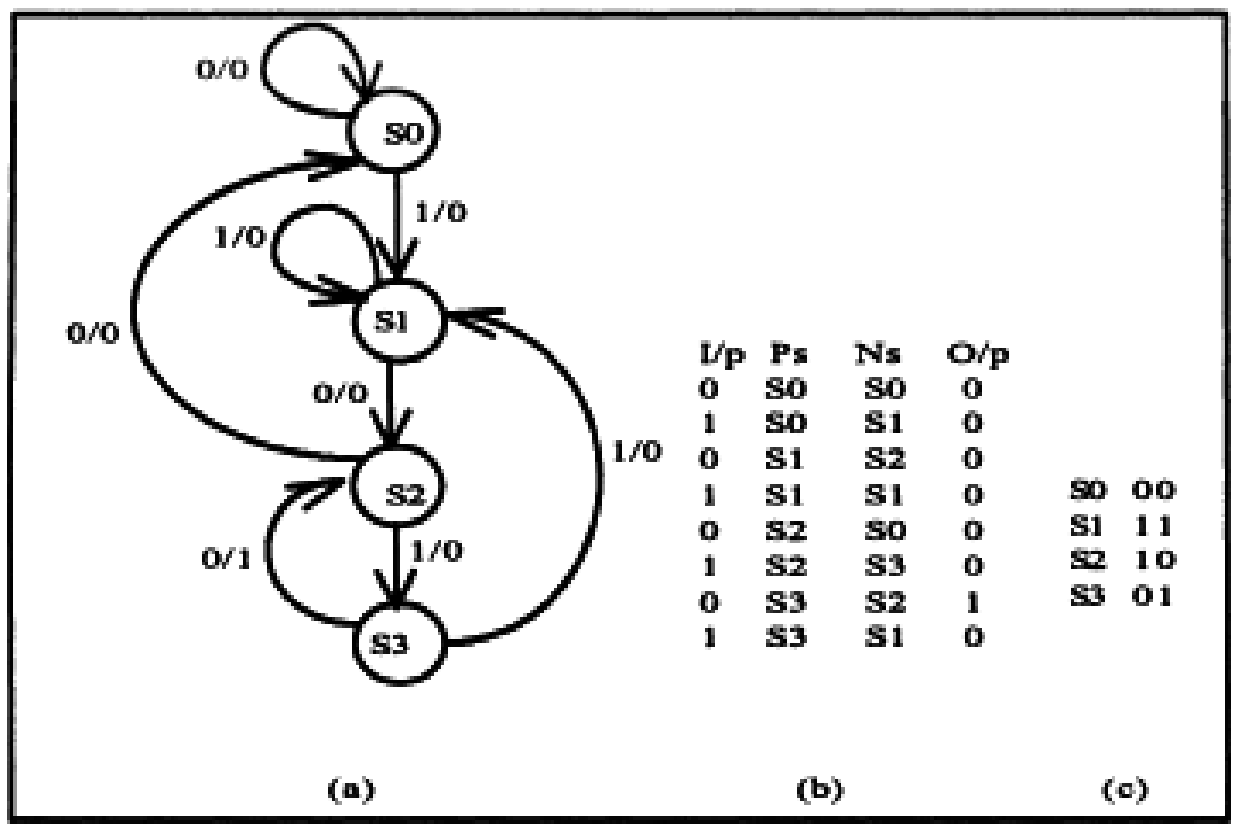

Figure 7. 1010 Detector (a) state transition graph, (b) state transition table, (c) state assignment[10].

This circuit includes two gates in subcircuit A and three gates in subcircuit C and there is not any gate in subcircuit B. The results that have been achieved by proposed method in compare with [12] have been shown in Table 1. In this circuit, maximum number of the generations for evaluation of subcircuit A was 4230 generations, for subcircuit B was 1300 generations and for subcircuit C was 5200 generations. We attained above results after 20 runs. In comparison with the method was presented in [12], our method uses the less gates, less generations, and the less times of evaluation to get 100% functionality. Also optimization decreases search space for GA by evolution of combinational parts of sequential circuit separately.

| i/p | Ps | Ns | o/p | | STT of the circuit | STT of subcircuit A | STT of subcircuit B | STT of subcircuit C |
|---|---|---|---|---|---|---|---|---|
| 0 | S0 | S0 | 0 | | .i 3 | .i 3 | .i 3 | .i 3 |
| 1 | S0 | S1 | 0 | | .o 3 | .o 1 | .o 1 | .o 1 |
| 0 | S1 | S2 | 0 | | .p 8 | .p 8 | .p 8 | .p 8 |
| 1 | S1 | S1 | 0 | S0=00 | 000 000 | 000 0 | 000 0 | 000 0 |
| 0 | S2 | S0 | 0 | S1=11 | 100 110 | 100 1 | 100 1 | 100 0 |
| 1 | S2 | S3 | 0 | S2=10 | 011 100 | 011 1 | 011 0 | 011 0 |
| 0 | S3 | S2 | 1 | S3=01 | 111 110 | 111 1 | 111 1 | 111 0 |
| 1 | S3 | S1 | 0 | | 010 000 | 010 0 | 010 0 | 010 0 |
| | | | | | 110 010 | 110 0 | 110 1 | 110 0 |
| i/p inputs | | | | | 001 101 | 001 1 | 001 0 | 001 1 |
| o/p outputs | | | | | 101 110 | 101 1 | 101 1 | 101 0 |
| Ps Present state | | | | | | | | |
| Ns Next state | | | | Step1 | Step2 | | Step3 | |

Figure 8. Process of STT of 1010 sequential circuit where .i input=input+present state bits, .o defined the number of outputs calculated, outputs of subcircuit A and B =next states of DFFs and output of subcircuit C =primary output bits, .p is the number of product terms

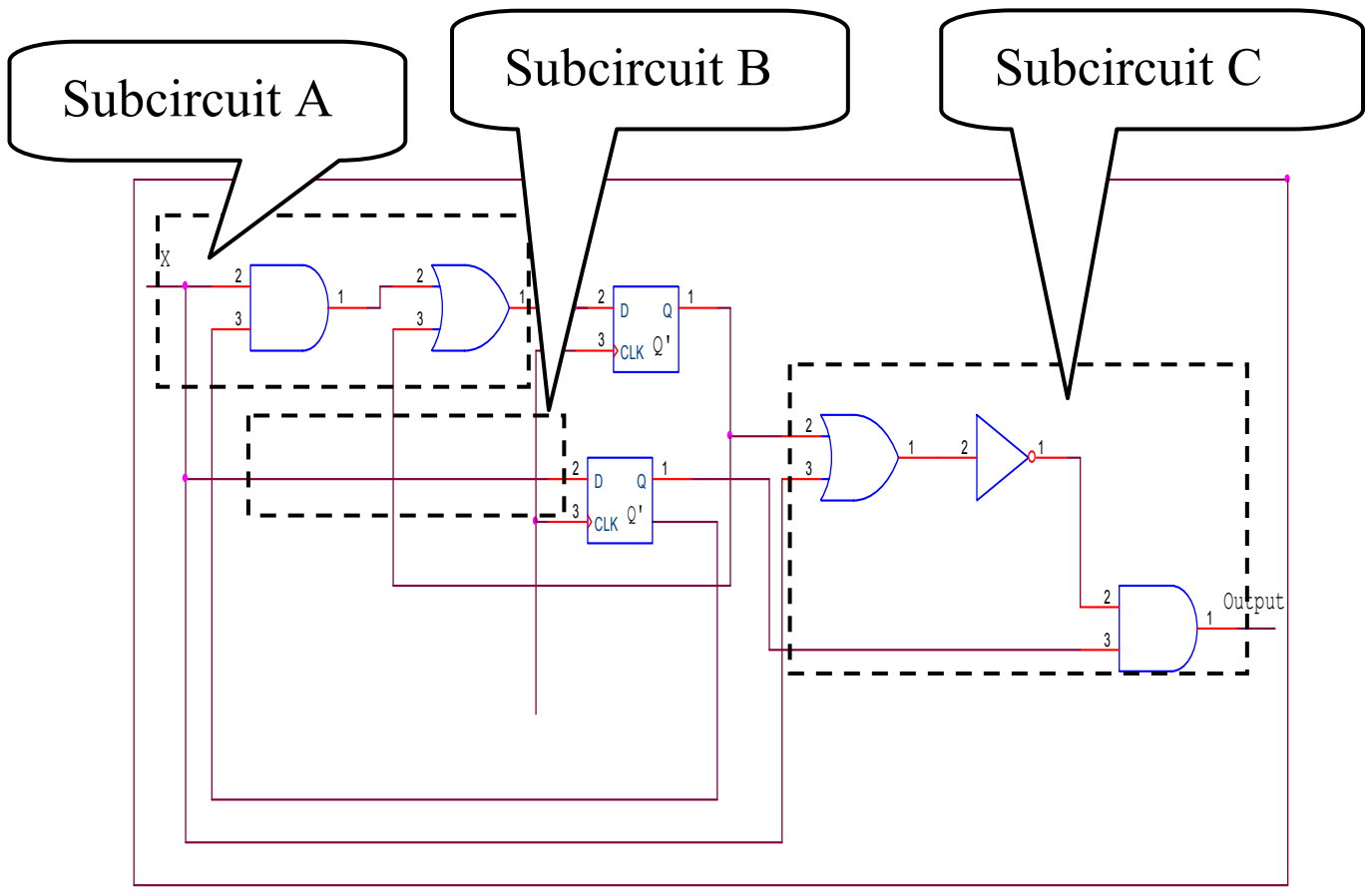

Figure 9. Evolved optimal circuit solution for 1010 detector.

TABLE I. SOLUTION OBTAINED FOR 1010 DETECTOR BASED ON FIG.9

| Proposal approach | Almaini [12] |
|---|---|
| $D_A$=XB'+A | $D_A$=X'A'B+X'AB'+XAB |
| $D_B$=X | $D_B$=A'B+AB'+XB' |
| Z=X'A'B | Z=X'AB' |
| Subcircuits of A,B=2 | Subcircuits of A,B=12 |
| Subcircuit C=3 | Subcircuit C=2 |

### *b) Sequential detector with 6 states*

We experiment another sequential detector in this section. This circuit has six states and uses three DFFs. State transition graph of this circuit has been shown in Fig. 10. We evolved this circuit similar to previous experiment. Fig. 11 depicts evolved circuit. In this circuit, subcircuit A has one gate, subcircuit B has five gates, subcircuit C has one gate and there is not any gate in subcircuit D.

In this experiment, the maximum number of the generations for evaluation of subcircuit A was 6120 generations, for subcircuit B was 10000 generations, for subcircuit C was 8310 generations and for subcircuit D was 8015 generations. We attained these results after 50 runs.

Table 2 compares our method with manual method and proposed method in [10]. The solution obtained by manual method, uses almost 2 times more gates than the circuit created by our method, and the method solution reported in [10] uses one gate more than our method. Maximum number of generations in [10] is 50000 generations, but in our method is 10000 generations.

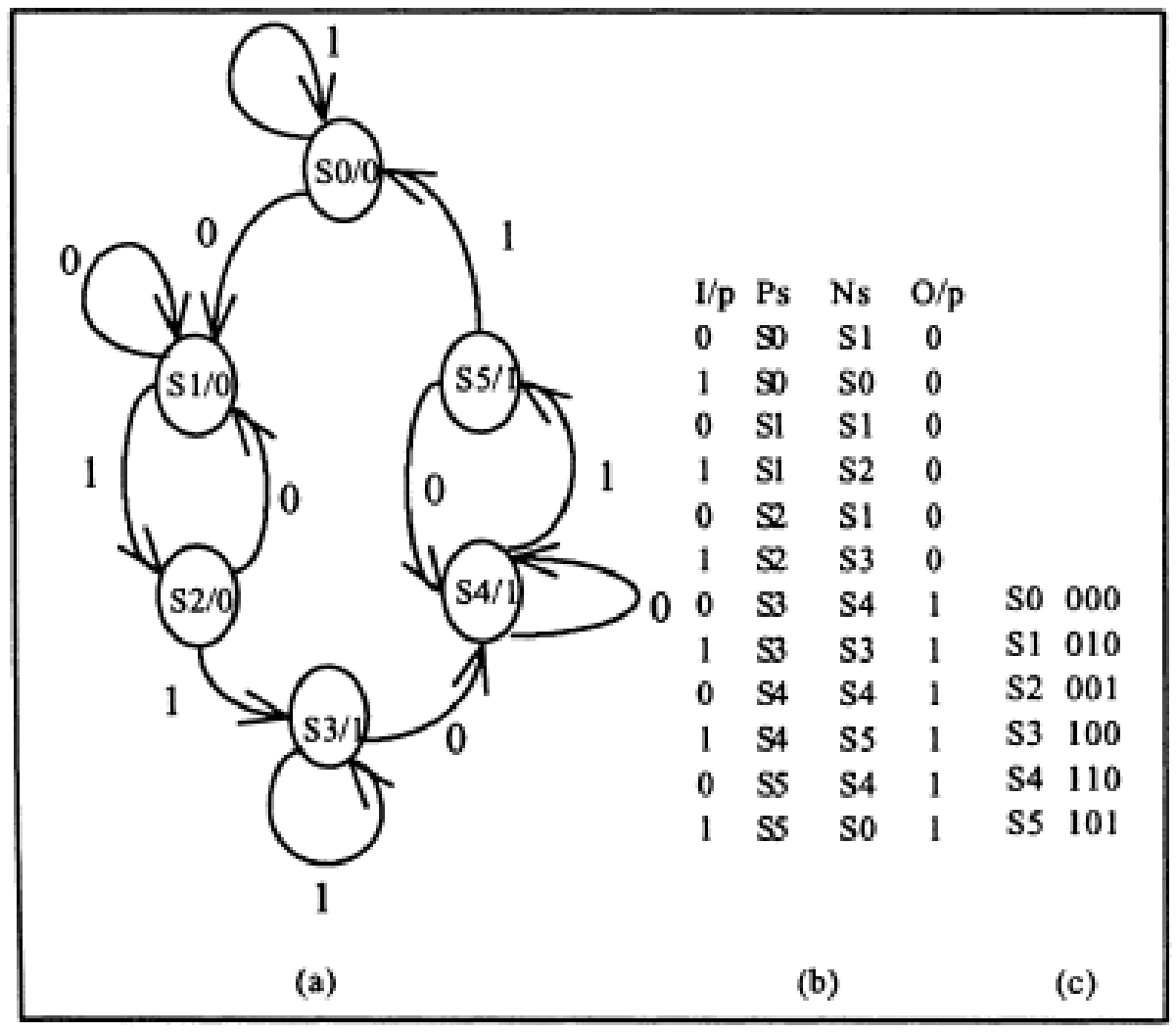

Figure 10. Sequential Detector (a) state transition graph, b state transition table, (c) state assignment [10].

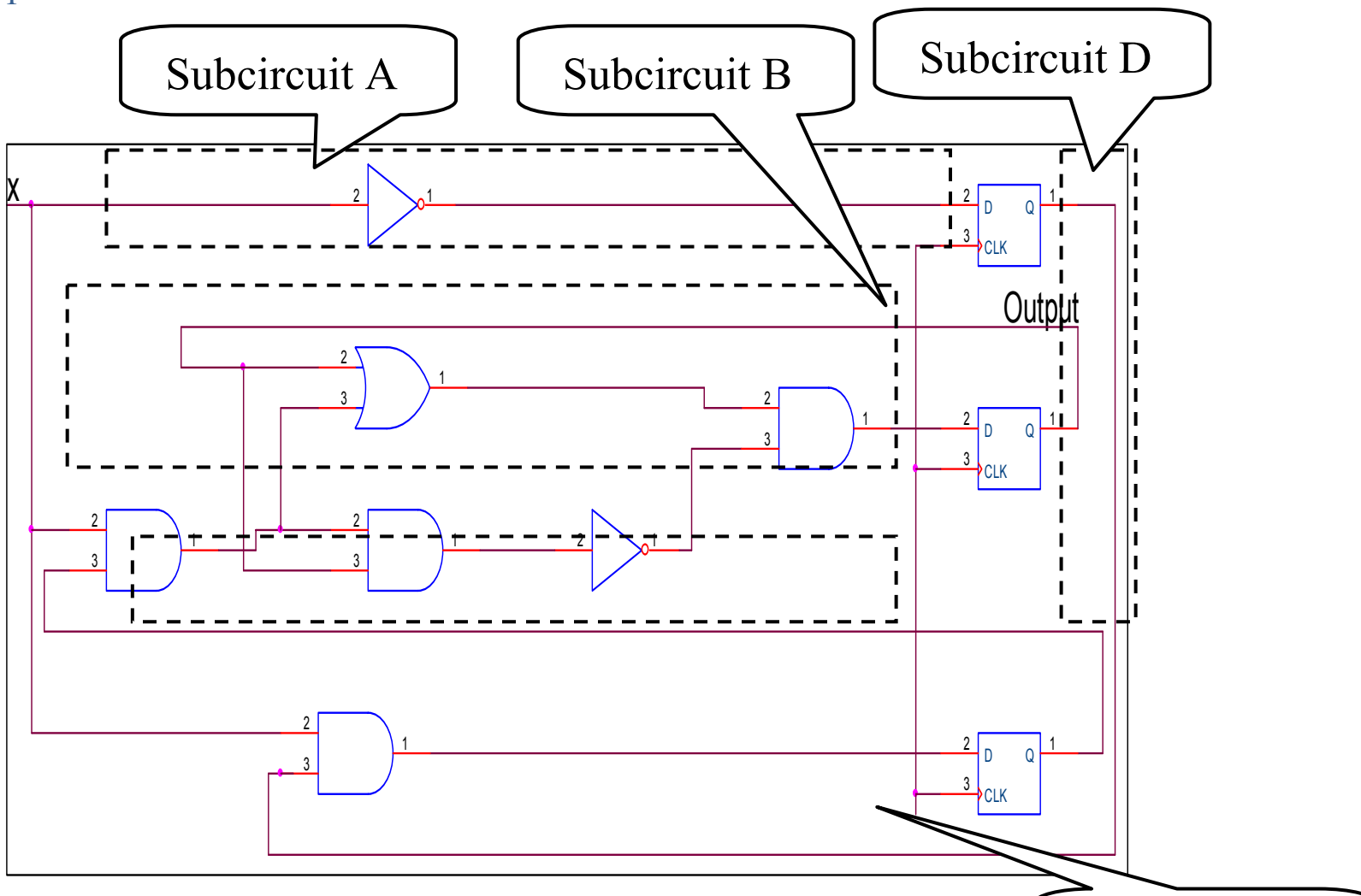

Figure 11. Evolved optimal circuit solution for sequential detector.

TABLE II. SOLUTION OBTAINED FOR SEQUENTIAL DETECTOR

| Proposed method | T.kalganova[10] | Manual method |
|---|---|---|
| $D_A$=XB | $D_A$ =XB | $D_A$=AC'+AX'+BCX' |
| $D_B$=X' | $D_B$=X' | $D_B$=BX+A'CX |
| $D_C$=(XAC)'(C+XA) | Dc=XAC'+X'C+A'C | $D_C$=BX+A'C'X'+A'B'X'+AC' X |
| Z=C | Z=C | Z=A+BC |
| Subcircuits of A,B,C=7 | Subcircuits of A,B,C=8 | Subcircuits of A,B,C=17 |
| Subcircuit D=0 | Subcircuit D=0 | Subcircuit D=2 |

## 8. Conclusions

This paper, has presented a method to design and optimize the synchronous sequential circuits. In this method, we have separated combinational parts and DFFs of sequential circuit and evolved them separately. This method decreased search space in GA and increased the speed of evolution. In comparison of our method with other methods, our method can design sequential logic circuits better than them and need to less time for evaluating. For future works it can be considered the evolution of the large scale sequential circuits by using proposed method that is applying more in industry.

**Parisa Soleimani** is MS candidate in electronic engineering from Central Tehran Branch of Islamic Azad University. Her research interest includes Evolvable Hardware, digital signal processing.

**Reza Sabbaghi-Nadooshan** received the B.S. and M.S. degree in electrical engineering from the Science and Technology University, Tehran, Iran, in 1991 and 1994 and the Ph.D. degree in electrical engineering from the Science and Research Branch, Islamic Azad University, Tehran, Iran in 2010. From 1998 he became faculty member of Department of Electronics in Central Tehran branch, Islamic Azad University, Tehran, Iran. His research interests include interconnection networks, Networks-on-Chips, Hardware design and embedded systems.

**Sattar Mirzakuchaki** received the BS in Electrical Engineering from the University of Mississippi in 1989, and the MS and PhD in Electrical Engineering from the University of Missouri-Columbia, in 1991 and 1996, respectively. He has been a faculty member of the College of Electrical Engineering at the Iran University of Science and Technology, Tehran, since 1996. His current research interests include characterization of semiconductor devices and design of VLSI circuits. Dr. Mirzakuchaki is a member of IEEE and IET (formerly IEE) and a Chartered Engineer.

**Mahdi Bagheri** received the PhD in Electrical Engineering from the Iran University of Science and Technology, Tehran, in 2011.